**Acquisition of Recursive Possessives and Recursive Locatives in Mandarin**

Chenxi Fu[a], Xiaoyi Wang[a], Ziman Zhuang[b] ,Caimei Yang[a]*

*a School of Foreign Languages, Soochow University, b Ulink High School of Suzhou Industrial Park*

***Corresponding author**: E-mail address:cmyang@suda.edu.cn(C. Yang)

**Abstract:** As recursion has been underlying any linguistic work for the last 60 years, the acquisition of recursive structures by children during language learning has become a focal point of inquiry. This study delves into the developmental trajectory of Mandarin-speaking children's acquisition of recursive possessives and locatives, assessing the impact of structural diversity on language acquisition. The research contrasts the comprehension of two-level recursive structures among children aged 3 to 7 years, employing answering question while seeing a picture task to elicit responses. The findings indicate that children do not attain adult-like proficiency in two-level recursion until the age of 6, and there exists a notable asymmetry in the acquisition of recursive possessives versus locatives. These results underscore the primacy of structural complexity and cognitive factors in the acquisition process, enhancing our comprehension of the cognitive foundations of language development and the pivotal role of recursion in child language acquisition.

## 1. Introduction

Language serves as the cornerstone of human communication, and its complexity stems from the diversity and recursive nature of its structure. Chomsky (1957) was instrumental in introducing the concept of recursion into the study of natural language. He posited that the grammar underlying human natural language consists of a finite set of recursive rules, which enable the generation of an infinite number of linguistic expressions. As Corballis (2014) noted, the assertion that recursion is the essence of natural language has been a recurring theme in Chomsky's work since his seminal 1957 publication, *Syntactic Structure*s. This theme was further elaborated in Hauser et al. (2002), who proposed that the faculty of language in the narrow sense encompasses only recursion, the sole uniquely human component of this linguistic capacity. This proposition has been succinctly encapsulated as the “recursion-only hypothesis” by Jackendoff and Pinker (2005: 212), underscoring the pivotal role of recursion in the field of linguistics.

Despite the absence of a universally accepted definition of (linguistic) recursion in the extant literature, the majority of studies focus on *category recursion*. This is characterized as the “embedding of a category within another instance of the same category.” For example, Martins and Fitch (2014) contended that recursion is utilized to describe the process whereby a constituent of a specific category is embedded within another constituent of the identical category. This "embedding" mechanism inherently gives rise to hierarchical structures that exhibit analogous properties across varying levels of embedding. Consequently, the attribute of "self-similarity" emerges as a distinctive hallmark of recursive structures.

To elucidate this concept, they cited the compound noun *[[sports] center]*, which adheres to the structural pattern [[[A]A]…]. This exemplifies recursion, as a noun phrase (NP) is nested within another NP. In contrast, a sentence comprising a noun followed by a verb, such as *[[things] happen]*, which conforms to the structural configuration [[[A]B]…], is deemed non-recursive. This is because a constituent of a particular category is not embedded within a constituent of the same category in such instances.

Under such a theoretical background, in order to understand the development of linguistic recursion among children, almost all acquisition practitioners (Matthei, 1982; Bryant, 2006; Hollebrandse et al., 2008; Limbach & Adone, 2010; Kinsella, 2010; Pérez-Leroux et al., 2012; Roeper, 2011; Hollebrandse & Roeper, 2014; Terunuma et al., 2017; Pérez-Leroux et al., 2018; Li et al., 2020, Bleotu & Roeper, 2021, 2022; Foucault et al. 2022, among others) involve *category recursion* sequences such as recursive possessives (e.g., *Elmo's sister's ball*), recursive prepositional phrases (e.g., *the baby with the woman with the flowers)*, recursive locatives (e.g., *tsukue-no osara-no ringo* 'Lit: table-Locative plate-Locative apple; an apple on the plate on the table'), recursive relative clauses (e.g., *the lion that is next to the bear that is next to the zebra*), and recursive complements (e.g., *I think you said they gonna be warm*). Among them, how children acquire the recursive possessives or recursive locatives is a popular problem. However, the conclusions were varied, and few compared the acquisition of recursive possessives with recursive locatives, thus shedding light on the influence of structural diversity. Looking for structures or constructions that cause difficulties in children, manifested often as avoidance, does not explain the reason for the complexity, which must be assessed with independent metrics. Thus, this study tries to investigate the acquisitive pathway in Mandarin-speaking children considering these two recursive structures.

## 2. Definition of recursion

The recursive rule “α→<α W>” is potentially left-branching, or left-expanding. In this rule, the non-terminal symbol α, which is enclosed in brackets, is positioned to the left of the terminal symbol W. It is important to note that Latin letters denote non-terminal symbols, while capital letters represent terminal symbols. Similarly, the recursive rule “β→<β M>” exhibits left-branching characteristics. Furthermore, rules such as “α→<α β>” and “β→<β α>” can be consolidated into the rule presented in (1a). This combined rule can then be integrated with (1b) to yield the recursive rule in (1c). Rules like the one in (1c) are potentially two-directional, encompassing both left and right-branching tail-recursion, as discussed by Rohrmeier et al. (2014).

(1a) α→<α <β α>>        (1b) β→M        (1c) α→<α <M α>>

As illustrated in (2a), the left-tail-recursion rule “N→N N” is applied consecutively twice through the sequence “N→[N N] N” (not “N→N [N N]”), with terminal symbols utilized to generate two-level tail-recursion sequences, such as in the example of “*wine bottle opener*”. In this diagram, circles symbolize recursive rules that are repeatedly employed, and arrows indicate the direction of expansion in a top-down manner.

(2) a.

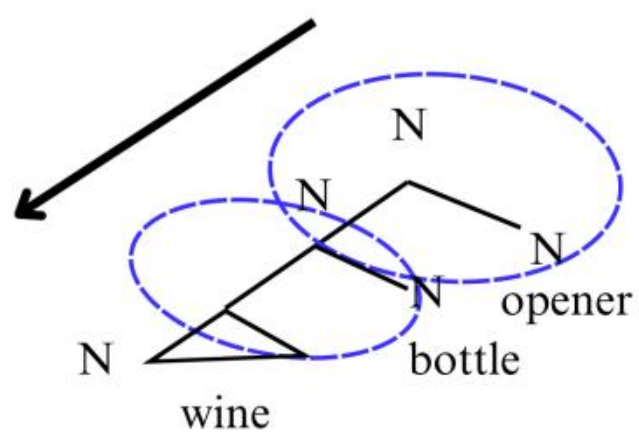

Also, as shown in (3a) below, the potentially left and right tail-recursion rule “N→N Poss N” (which results from the combination of the two rules “N→Poss N” and “Poss→N Poss”) is used twice consecutively by way of “N→[N Poss N] Poss N”, which is further filled with terminal words to generate two-level tail-recursion sequences such as *Elmo's sister's ball* and *Xiaoming de mama de yifu* 'the clothes of Xiaoming's mother'. Similarly, as shown in (3b), the tail-recursion rule "N→N Loc N" (which results from the combination of "N→Loc N" and "Loc→N Loc") is used twice consecutively by way of "N→[N Loc N] Loc N", which is further filled with terminals to generate two-level tail-recursion sequences such as *zhuozi-shang de diezi-shang de pingguo* 'an apple on the plate on the table'.

(3) a.

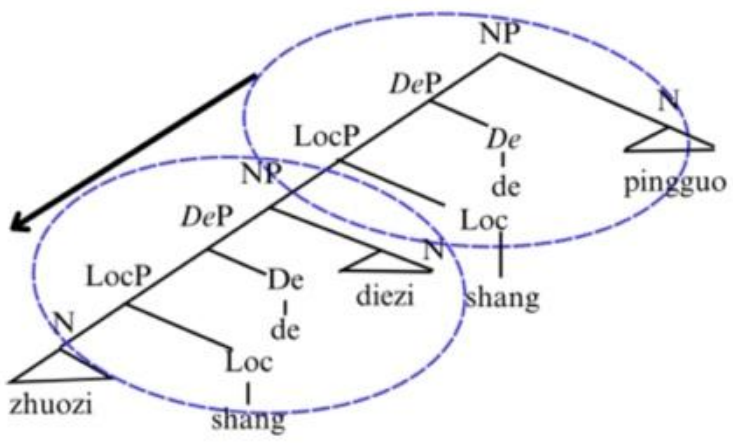

b.

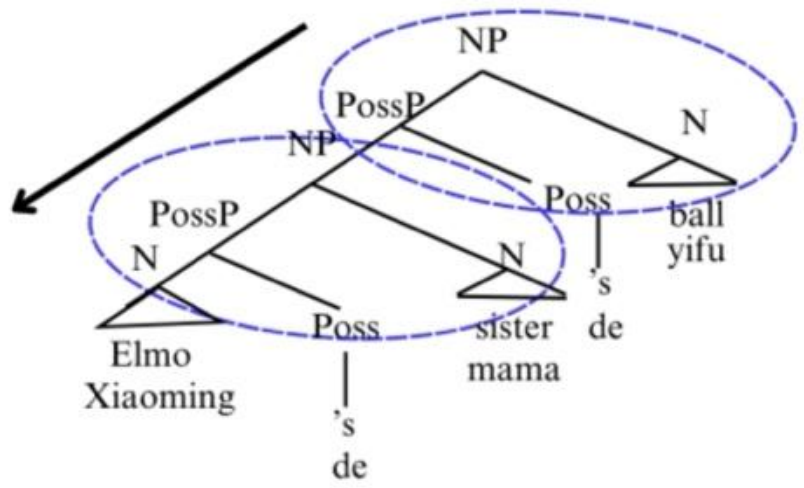

## 3. Literature Review

### 3.1 Previous acquisition studies on recursive possessives and locative

Figure 1 presents a detailed synthesis of key research studies examining the acquisition of recursion in children across various languages, including English, Japanese, Chinese, Tamil, Hungarian-Romanian bilingual, French, German, Persian, Romanian, and Spanish. The figure encapsulates the research purpose, age of participants, and acquisition ratios of locative and possessive recursion for each study.

*Figure 1. Summary of Key Research.*

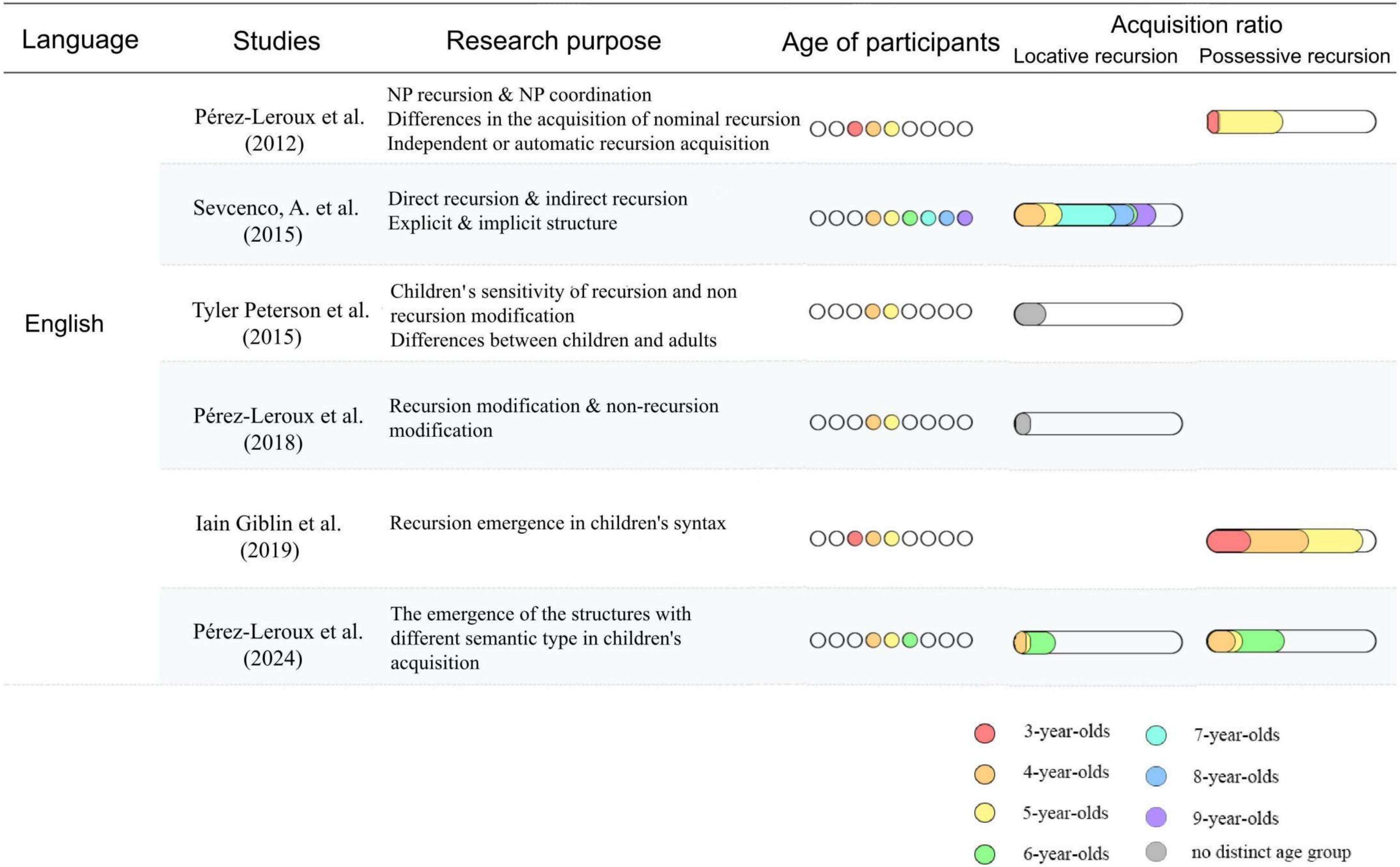

*Figure 1. Summary of Key Research.*

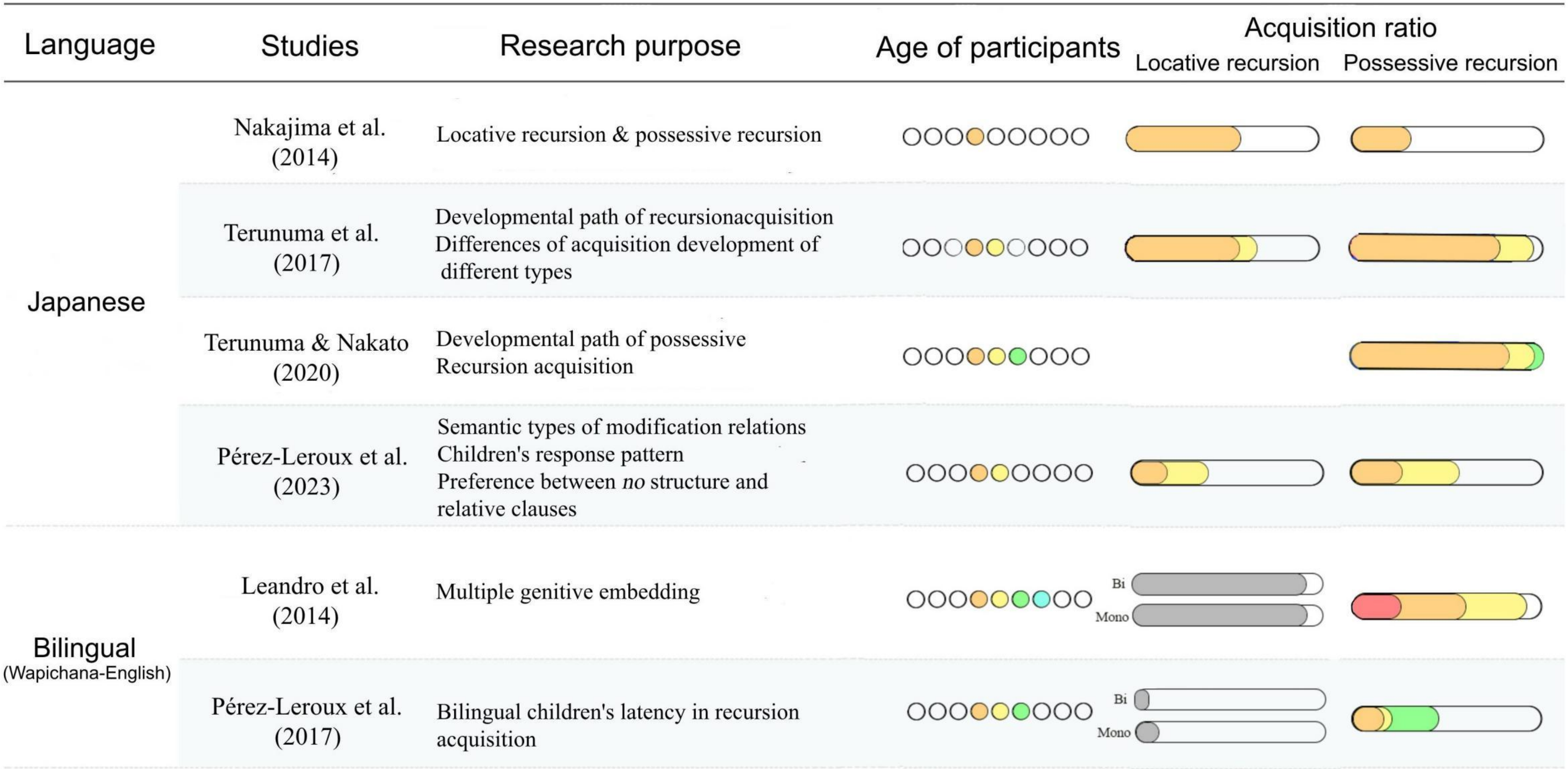

*Figure 1. Summary of Key Research.*

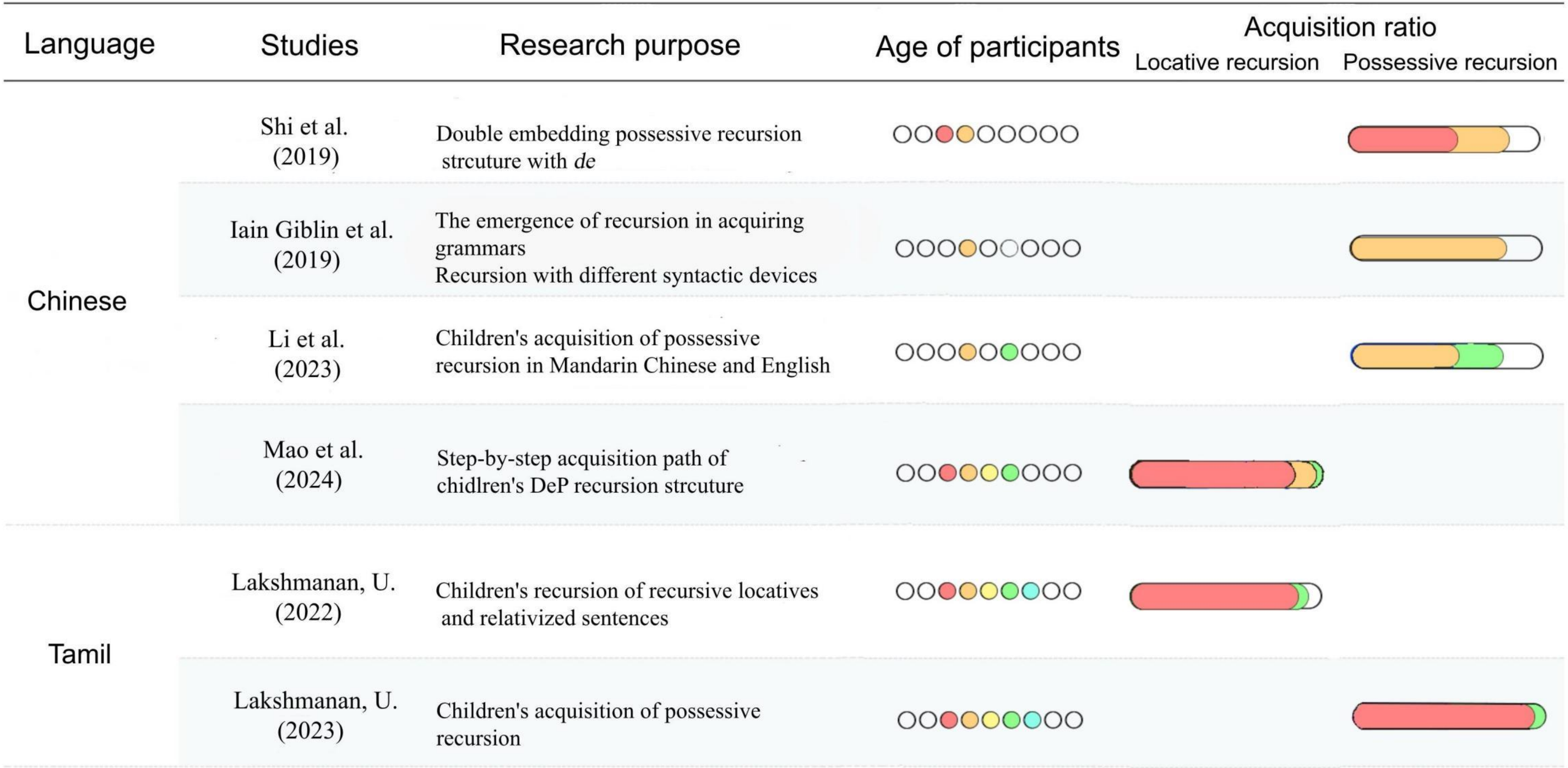

*Figure 1. Summary of Key Research.*

| Language | Studies | Research purpose | Age of participants | Acquisition ratio: Locative recursion | Acquisition ratio: Possessive recursion |
|---|---|---|---|---|---|
| Bilingual (Hungarian-Romanian) | Larisa Avram et al. (2020) | Recursion acquisition comparison bewyeen L2 and L1 | | Bi / Mono | |
| French | Roberge et al. (2018) | Performance and difficulty in recursion acquisition | | | |
| German | Pérez-Leroux et al. (2022) | The influence of structural variants on recursion acquisition | | | |
| Hungarian | Toth, A. (2017) | Children's explaining of recursion The role of functional head | Pre-school children Grade 2 children | | |
| Persian | Hossein et al. (2023) | 3 steps for recursion acquisition | | | |
| Romanian | Adina Camelia Bleotu (2020) | Conjunction & recursion | | | |
| Spanish | Pérez-Leroux, A. (2022) | Developmental timeline of recursive nominal modi cation in Spanish | | | |

*Figure 1. Summary of Key Research.*

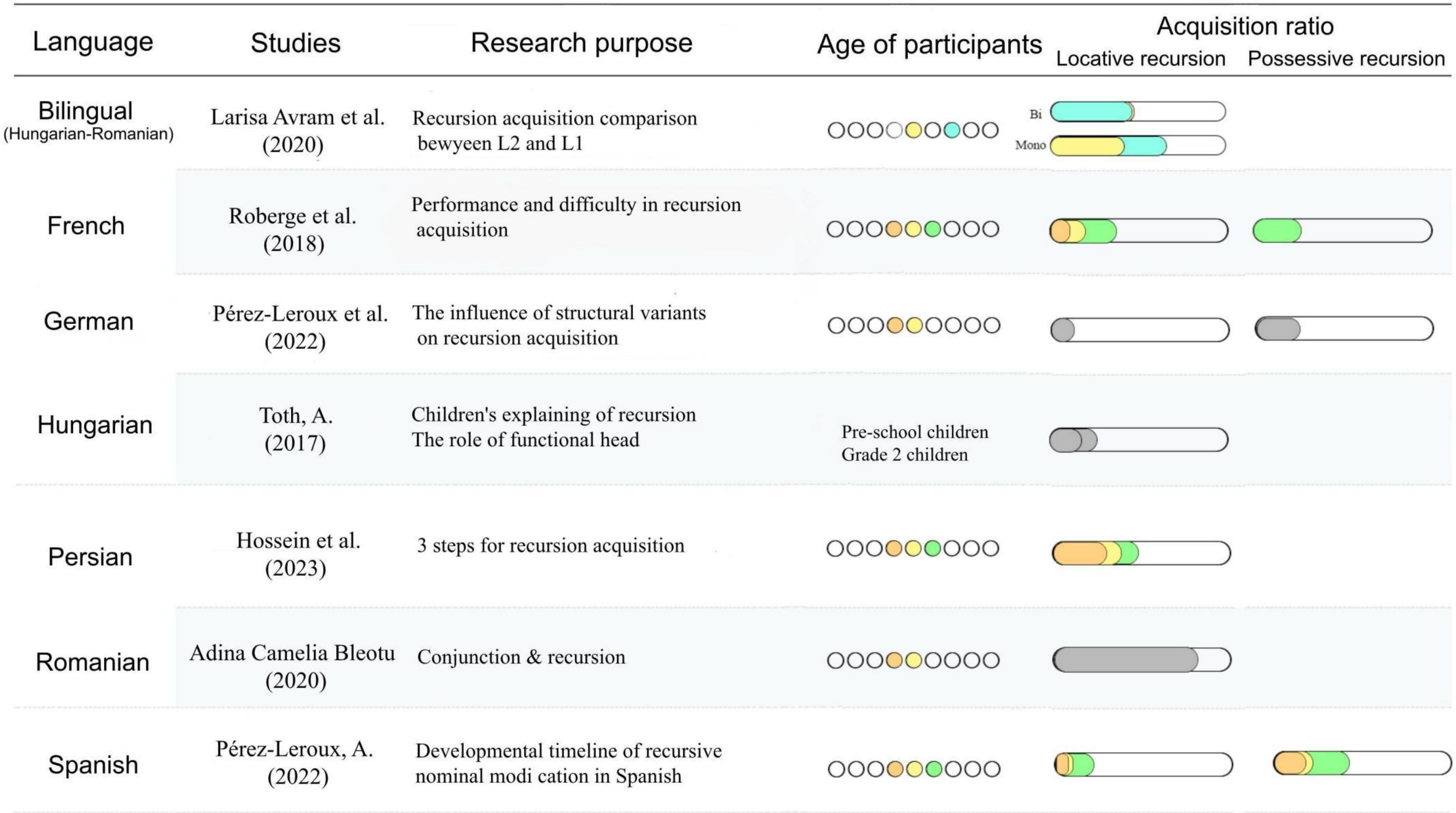

The data presented in Figure 1 indicate that the acquisition of tail-recursion sequences is a phenomenon observed across various languages, albeit with variations in the timing of acquisition and an asymmetry between possessive and locative recursion sequences. The majority of participants in these studies are between the ages of three and six, suggesting that this is a critical period for the development of recursion in children's language acquisition. The findings from these studies contribute to a deeper understanding of the developmental stages and patterns in children's acquisition of recursive language structures across different linguistic contexts.

More specifically, there is a lack of investigation on the studies revolving around Mandarin tail-recursion sequences. Zhou Peng’s team has done great studies in possessive recursion. Shi et al, (2019) recruited 30 monolingual Mandarin children under 4 years old to investigate the production of 2-level possessive recursion (e.g., *hǎidào de qīngwā de bǐnggān*; pirate's frog’s cookies’) and found that 4-year-olds can successfully produce two-level recursive possessives. In Iain Giblin’s (2019) second study, the Mandarin children’s acquisition of 2-level possessive recursion was researched at the age of four by watching the picture while hearing the story and then making choices. Apart from these, Li et al. (2020) did a comprehensive experimental analysis of Mandarin children’s acquisition of 2- and 3-level possessive recursion structure, in which researchers described the possessive relationship in the picture with recursion structure to children. Then children were elicited to produce corresponding possessive relationships(e.g., *jīqìrén de shīzi de shé*; robot’s lion’s snake’) and found that the accuracy of children under 6 is lower than 70%. Besides, Author et al.(2023) conducted a "picture-elicited-minimal-pair production" experiment on three types of recursive relative clauses, and found that children acquire 2-level tail-recursion RCs (e.g., *gǒu yǎo de māo dǎ de hóu*; the cat that bit by the dog hit the monkey) at 7, showing that children acquire tail-recursion at late age. For the locatives, only Mao et al.(2024) conducted an experiment on 84 children between 3 and 6, which demonstrates that children can acquire 2-level recursive locatives (*huāyuán lǐ de yǐzi shàng de māo*; a cat on the chair in the garden) at the age of four. However, the experiment utilized a pointing task that required children to indicate their responses through gestures. Given that children's pointing can be capricious, the reliability of the findings is heavily contingent upon their ability to follow simple directives, which may not sufficiently capture the nuances of more sophisticated language comprehension and production mechanisms. Furthermore, the experimental stimuli, particularly the test images, were not always designed with sufficient clarity or intuitiveness. The proximity of the target object to distractors could lead to children selecting an incorrect object due to ambiguous visual cues, or there is a potential for researchers to misinterpret children's responses owing to varying perspectives. These factors could introduce a degree of inaccuracy into the assessment of children's understanding of recursive syntactic structures.

Research into the language acquisition of children with non-English mother tongues has yielded analogous findings: children are capable of accurately comprehending and producing two-level and higher-level recursive structures at an advanced stage of linguistic development. Terunuma et al. (2017) utilized a "picture-listening question" approach to assess the understanding of recursive possessives and locatives among Japanese children aged 4 to 5 years. Their findings indicated that the responses of 4- and 5-year-olds to 2-POSS and 2-LOC sentences were notably akin to those of adults, with children generally providing fewer adult-like responses to recursive locatives compared to recursive possessives at each level of recursion. Fujimori (2010) presented subjects with pictures and narratives, followed by questions, demonstrating that Japanese-speaking children can acquire two-level possessive recursion by the age of four. Terunuma and Nakato (2018), modifying Roeper's (2011) design, discovered that 79.2% of 4-year-old Japanese children could produce responses that reflected an adult-like mastery of two-level recursive possessives. Consistent with these findings, studies on recursive locatives (Nakajima et al., 2014; Sevcenco et al., 2015; Terunuma et al., 2017; Roberge et al., 2018; Pérez-Leroux et al., 2017, 2023, 2024) have observed that children typically acquire this structure during the later phases of language development. In contrast, Pérez-Leroux et al. (2012) noted that while children readily produce coordinated nominals, the generation of recursively complex nominals presents a significant challenge, prompting them to eschew recursive structures in favor of alternative response types. This tendency was also observed by Sevcenco et al. (2015). Pérez-Leroux et al. (2012) hypothesize that coordination may bypass the merging process characteristic of recursion, potentially emerging from a more primitive form of concatenation or aggregation, which might be developmentally prior in the child's linguistic inventory.

Some research designs pose problems. In experimental paradigms that involve intricate narrative contexts and complex character relationships, participants' responses are often contingent upon their memory capacity (Shi Jiawei, 2019). The study by Terunuma & Nakato (2013) failed to meticulously delineate the variables under investigation from those that potentially confound children's linguistic acquisition. The presence of an excessive number of distracting elements in the visual stimuli further complicated the task, impeding Japanese children's ability to make informed decisions. Subsequent refinements to the original experimental design by Terunuma et al. (2017) underscored the validity of this critique: each level of the test targeted the assessment of the corresponding recursive structure, with the test graph illustrating the incidence of children's omissions of single or multiple related phrase structures during their comprehension of recursion. This approach effectively mitigated the cognitive demands on the participants. The present investigation adopts a modified approach from Terunuma et al. (2017), eschewing the narrative element to mitigate the cognitive load on children's memory. The visual stimuli utilized in this study feature characters and objects that are ubiquitous in everyday contexts. These elements were introduced to the participants prior to the experimental phase to preclude any recognition-related impediments for the children.

**3.2 Explanation of Asymmetry Between Structures**

Scholars have proposed multiple perspectives to elucidate the dimensional differences in language complexity as observed in acquisition studies. Bejar et al. (2020) have concluded that linguistic complexity should not be solely defined by the iterative applications of Merge. Their analysis is grounded in four key perspectives. Firstly, they identify an asymmetry between coordination and modification structures. Children exhibit an ease in acquiring coordinating structures relative to embedded ones, as demonstrated by Pérez-Leroux et al. (2012). This disparity suggests a differential acquisitional challenge for children, despite the syntactic similarities between these structures. Secondly, there is an asymmetry between sequential and recursive PP modifications. Research by Pérez-Leroux et al. (2018) indicates that children encounter less difficulty with sequential modifications (e.g., "the plate under the table") compared to nested modifications (e.g., "the bird on the crocodile in the water"). This finding implies that even structurally analogous forms, recursive nesting may present a heightened complexity during the acquisition process. Thirdly, an asymmetry is noted in relative clause modification versus recursive PP modification. In their experiments, Pérez-Leroux et al. (2018) observed a propensity for both adults and children to utilize more complex relative clause structures in place of the target PP modifier, despite these structures not contributing additional informational content. This preference hints at a complexity that transcends syntactic intricacy, potentially relating to the semantic and pragmatic interface processing in language. Fourthly, qualitative disparities are observed between genitives and PPs. The complexity arising from different types of NP embeddings may not be linked to structural metrics. Roeper & Snyder's (2004) observations on the German possessive form highlight that the acquisition of a rule (e.g., possessive -s) is a distinct learning phase from the acquisition of rule iteration, which permits multiple embeddings.

Pérez-Leroux et al. (2024) corroborates the notion that recursion development is bounded by experiential factors and productivity, with its utilization contingent upon third-factor considerations, such as processing capabilities. These findings challenge the initial work of Pérez-Leroux et al. (2012), suggesting that neither children nor adults are impeded by the structurally distinct possessive -s.

In contrast to the aforementioned views, Pérez-Leroux et al. (2022), through their investigation into the impact of structural diversity on recursion acquisition, propose that such diversity does not impede children's mastery of recursive expressions within a specific domain. While structural complexity may influence the overall timing of the onset of recursive modification, it does not sufficiently account for performance variability across different domains or the specific choices children make in language use.

Building upon the modified methodology of Terunuma et al. (2017), the current study aspires to extend the research on the acquisition of two distinct types of recursion—possessives and locatives—to ascertain the developmental timeline during which children grasp these structures and the trajectory of their acquisition. Additionally, this investigation seeks to ascertain the extent to which structural diversity impacts the accessibility of recursion for children. To accomplish these objectives, the study addresses the following three pivotal questions:

1. When can Chinese children comprehend the two-level recursive possessive structure and recursive locative structure respectively?

2. What are the characteristics of Chinese children’s acquisition of recursive possessive and

recursive locative structure respectively? Are there any differences in the development of the two recursive structure?

3. What factors could affect acquisition order of the two structures?

## 4. Experiment

The following sections describe the design and methodological components of our study.

### 4.1 Participants

One hundred and twenty-seven children in ages between 3 and 7 years participated in the experiment. One participant did not complete the session and was removed from the analysis. They were divided into four age groups: 3-year-olds (N = 30, M = 3;08, SD=0.17, range = 3;05 – 3;11), 4-year-olds (N = 35, M = 4;06, SD=0.28, range = 4;00 – 4;11), 5-year-olds(N=28, M = 5;05, SD=0.27, range = 5;01 – 5;11) and 6-year-olds (N = 34, M = 6;08, SD=0.25, range = 6;00 – 6;10). The children were recruited from a Kindergarten in Fujian Province. All of them were monolingual speakers of Mandarin. No participant had any history of speaking or hearing difficulties or cognitive impairment. Gender and socioeconomic status were not considered. Thirty adults aged 21 to 53 (SD=9.13) also participated in the study serving as control group.

### 4.2 Materials and procedures

We adapted the "answering a question while seeing a picture" designed by Terunuma et al.(2017) without story, which relieve memory burden, to elicit the children's comprehension of recursive possessives and recursive locatives. The participants' task was to answer a question while seeing a picture. The task targeted understanding of two-level recursive possession and recursive location(1)-(2) with 10 nominal recursion questions (e.g. The second green car misses tires, is that right?) as fillers to avoid familiar effect. The experiment consists of two stages, pre-test and post-test. In the pre-test stage, the experimenter showed the subjects the objects and characters that would appear separately without involving possessive relationships, so as to prevent the subjects from not knowing the objects in the experiment topics during the experiment. There is no pre-test phase for adult.

(1) 2-POSS

*Jiějié-de tùzi-de qiú shì shénme yánsè?*

sister-Poss rabbit-Poss ball is what-Q color

What color is the girl's rabbit's ball?

(2) 2-LOC

*Yǐzi shàngde nàozhōng shàngde lǎoshǔ shì shénme yánsè?*

chair on-Loc clock on-Loc mouse is what-Q color

What color is the mouse on the clock on the chair?

In the post-test, the experimenter showed the test questions to the subjects via ppt on the computer. The test included pictures and the voice prompt of the question. The picture in the test contains a two-level recursive possessive or recursive locative relationship, and the questions contained in the test would ask questions containing the two-level recursive structure in the picture. The prompting voice of the test questions was made by AI, and each question was played twice, and the subjects were asked to answer after listening to the question twice. For example, for Figure 2, the question is "*Jiějié-de tùzi-de qiú shì shénme yánsè?* /What color is the girl's rabbit's ball?" and for Figure 3, the question is "*Yǐzi shàngde nàozhōng shàngde lǎoshǔ shì shénme yánsè?* /What color is the mouse on the clock on the chair?" After the question was played twice, the participants required to identity the colors of objects in the picture.

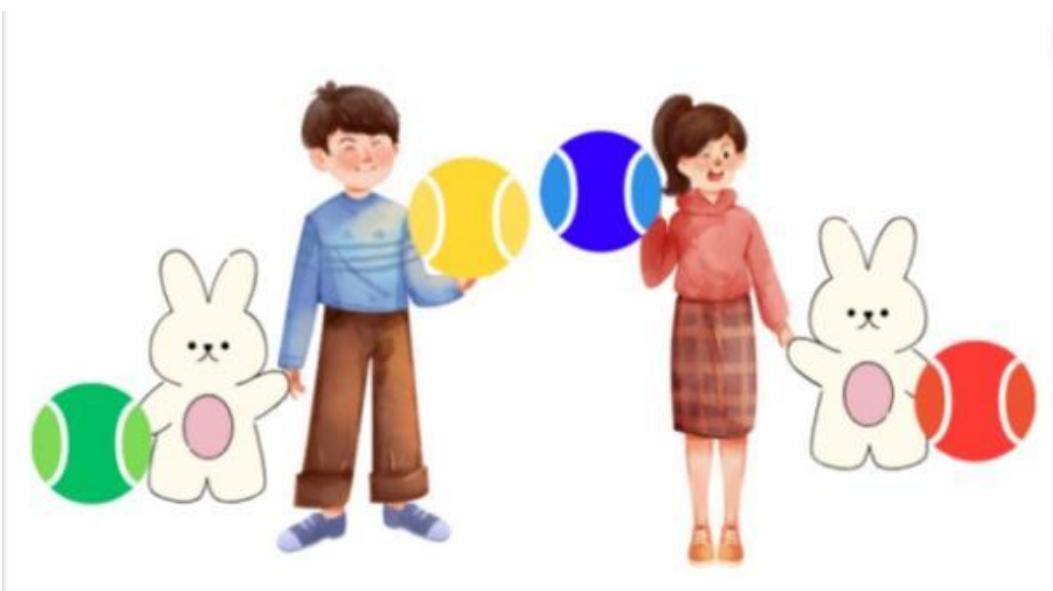

**Figure 2. Sample picture for 2-POSS.**

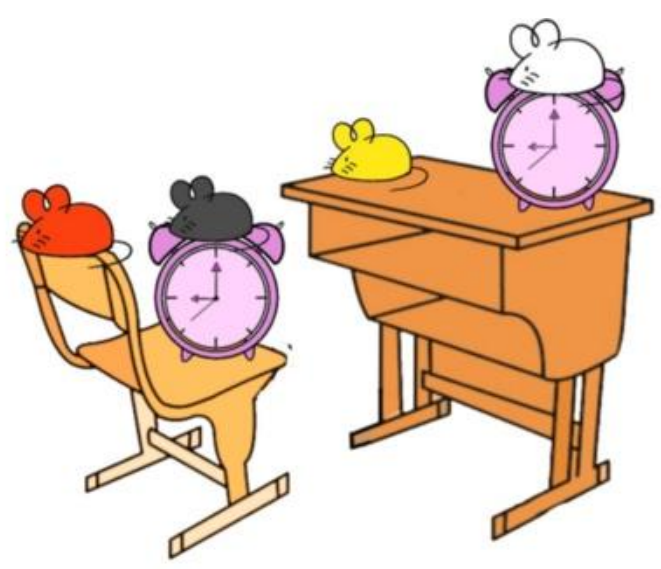

**Figure 3. Sample picture for 2-LOC.**

Since it has been reported that there is possibility that children often drop one of the recursive phrases or make conjunctive response in their comprehension of recursive structures (Gentile 2003; Limbach and Adone 2010), we took into consideration possible conjunctive and dropping interpretations of the target sentences when we design entities in the pictures. For both conjunctive reading and dropping reading, there are referents providing for all the possible interpretations different from recursive reading in the pictures. Specifically, in the test question of Figure 2, if the subject dropped the possessive structure "sister's" that is the first level, the corresponding answer should be "red and green"; If the subject drops the possessive structure of the second level, "rabbit's", the corresponding answer should be "blue"; If the two-level possessive structures were dropped, then the subject's answer should be "red, yellow, blue and green." The tests for the recursive locative structures were made in a similar way.

There were 4 questions with two-level possessive recursion and 5 questions with two-level locative recursion. These test were assigned to the participants in two sessions, in order and in reverse order. Half of the child participants started with sequential order and the other half with reverse order. The adult participants followed the same pattern. In the test phase, no matter what answer the subject answered and their accuracy, the experimenter gave positive feedback. The experimenter would not correct the subjects' answers if they made errors.

During the experiment, the answers of the subjects will be recorded or videotaped by mobile phone or camera, and the experimenters wrote down the answers of the subjects during the experiment. The experiment was conducted individually in a quiet and comfortable room for about 10 minutes for each child.

**4.3 Data Coding**

In the study, children's responses to the inquiry questions corresponding to each recursion type were meticulously transcribed and systematically categorized based on their accuracy and the nature of any errors. Upon the experiment's culmination, the researchers discerned a variability in the children's responses; some participants provided answers upon the initial posing of a question, whereas others either reiterated their previous responses or altered them when the question was reiterated. The experimental protocol incorporated a lenient coding strategy, wherein participants were awarded a single point for a correct response, irrespective of whether it was their first or second attempt. Conversely, participants received zero points for two consecutive incorrect responses. Among the erroneous responses, further classification was conducted into three categories: Conjunction (indicating the sequential interpretation of the two determiner phrases by the participants), Drop (denoting the omission of at least one genitive determiner phrase), and Other Interpretation (encompassing all other types of incorrect responses). Following the coding process, data analysis was conducted utilizing SPSS, and the ensuing experimental outcomes are delineated in the subsequent sections.

## 5. Results

### 5.1 Analysis of Targeted Answers

Table 1 and Figure 4 present the distribution of response types across various age groups for tests involving recursive possessive and locative constructions. Notably, the responses from the six-year-old children closely resembled those of adults. In contrast, the responses from the three-, four-, and five-year-old children significantly diverged from the adult responses for both types of recursion. Furthermore, it was observed that children consistently provided fewer responses that aligned with adult-like answers when dealing with recursive locatives as opposed to recursive possessives at each level of recursion.

| | Recursive Types | Recursion (Correct) | Conjunction (Error) | Drop (Error) | Other Interpretation (Error) |
|---|---|---|---|---|---|
| Adults(n=30) | PP | 99.17% | 0% | 0% | 0.83% |
| | LP | 98.67% | 0% | 0% | 1.43% |
| 6-year-old(n=34) | PP | 87.5% | 2.94% | 5.15% | 4.41% |
| | LP | 63.64% | 9.70% | 11.52% | 15.15% |
| 5-year-old(n=28) | PP | 75% | 11.61% | 5.36% | 8.04% |
| | LP | 48.28% | 10.34% | 8.28% | 33.10% |
| 4-year-old(n=34) | PP | 67.86% | 7.14% | 11.43% | 13.57% |
| | LP | 42.29% | 8.57% | 17.14% | 32.00% |
| 3-year-old(n=30) | PP | 65% | 5.04% | 14.28% | 15.97% |
| | LP | 46.00% | 2.67% | 20.67% | 30.67% |

**Table 1 Percentage of answers of recursive possession and recursive location.**

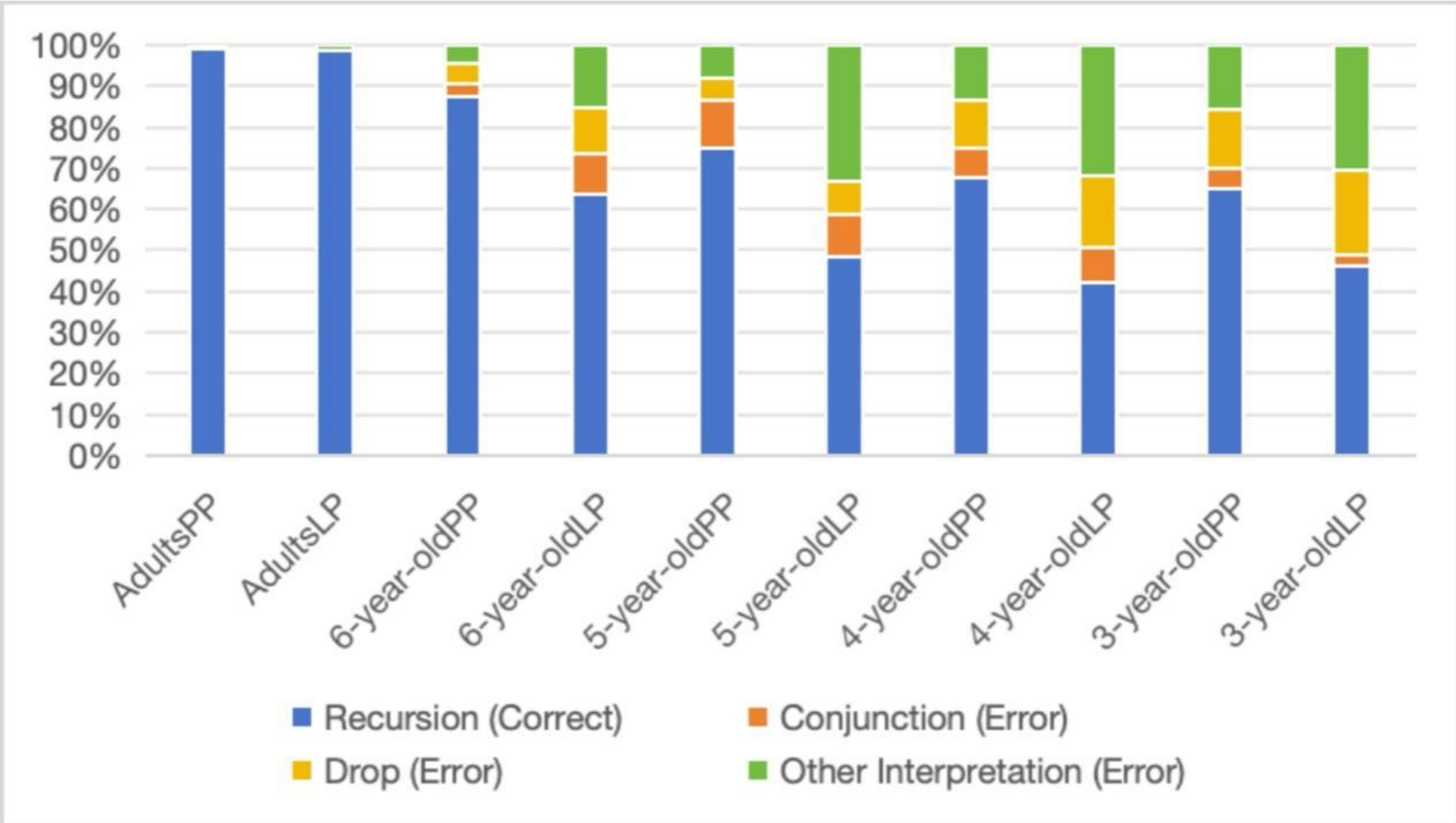

**Figure 4. Correct Percentage and Incorrect Percentage of Answers.**

The acquisition of recursive possessives and locatives by Chinese-speaking children is characterized by a gradual developmental trajectory. Figure 5, which is derived from the accuracy rates of the participants, illustrates the mean proportion of target responses elicited from different age groups for the two types of recursion. As depicted in Figure 5, there is a discernible upward trend in the accuracy of responses for both recursive possessives and locatives with increasing age, signifying a progressive mastery of these linguistic constructs. However, a distinction is observed between the acquisition patterns of recursive possession and locative structures. The data indicate that children's accuracy in producing recursive possessives consistently exceeds that of recursive locatives at corresponding age brackets, with the disparity in correctness rates reaching up to 25.57%. Notably, even at the age of 6-7 years, children's proficiency in handling recursive locatives remains significantly below the 75 percent threshold, whereas their performance on recursive possessives approximates adult levels. These findings suggest that the developmental pathways for recursive possessive and locative structures diverge within the child language acquisition process. The differential rates of acquisition highlight the distinct cognitive and linguistic challenges posed by these two grammatical phenomena, underscoring the need for further investigation into the factors that influence their acquisition.

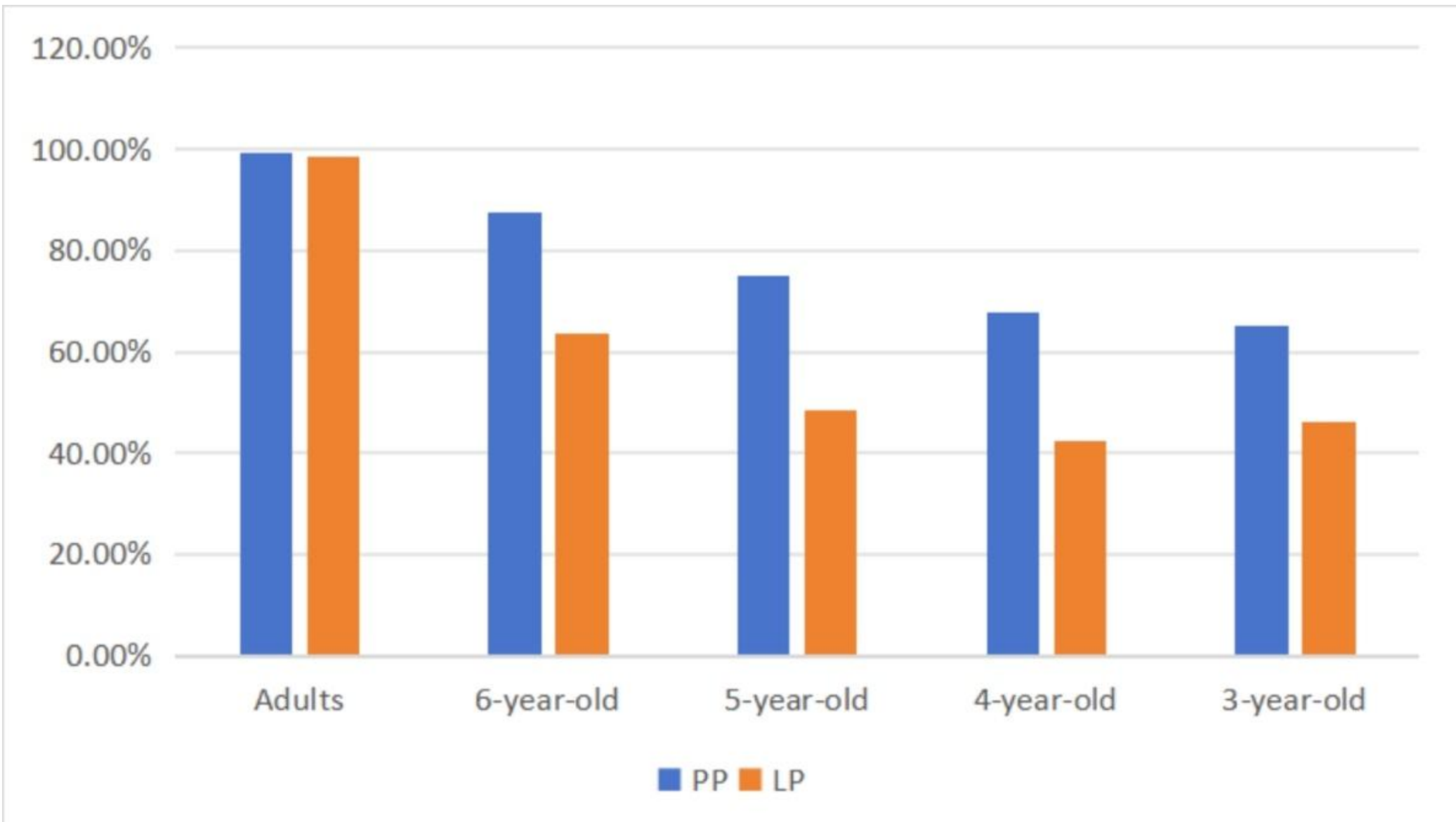

**Figure 5. Correct Percentage in Each Age Group.**

**5.2 Analysis of Untargeted Answers**

In the analysis of inaccurate responses, particular attention was given to the patterns of non-target interpretations, specifically focusing on conjunction and dropping errors.

Within the recursive possessive structure, the conjunction error accounted for 6.68% of the total non-target responses, while the drop error, characterized by the omission of at least one determiner phrase (DP), represented 9.06%. In the context of "drop" errors, participants often omit a possessive structure during comprehension, perceiving the sentence as a single-layer recursive construct. For instance, they might interpret structures such as "[NP [PossP [NP sister] [Poss 's]] [N ball]] / the sister's ball" (Figure 6) or "[NP [PossP [NP rabbit] [Poss 's]] [N ball]] / the rabbit's ball" (Figure 7) as simplified, single-level recursions. This error type was particularly pronounced in 3-year-olds, constituting 14.28 percent of their responses. However, as children age, the frequency of this error type diminishes significantly, dropping to merely 5.15 percent by the age of 6. Regarding "conjunction" errors, participants tend to sever the rightward two-level possessive recursive structure, uniting NP1 and NP2 to form a single modifier for NP3. This results in an interpretation that reduces the structure to a straightforward, first-level recursive sequence. An example of this misinterpretation would be the understanding of "[NP3 [Poss1 [NP1 sister] [Poss 's]] [Poss2 [NP2 rabbit] [Poss 's]] [N ball]] / the ball shared by the sister and the rabbit" (Figure 8) as a dual-possessive modifier. In the experiment, the incidence of such errors escalated from 5.04 percent in 3-year-olds, peaking at 7.14 percent in 5-year-olds—a rate considerably higher than that of omissions during the same period—before declining to their lowest rate of 2.94 percent by the age of 6.

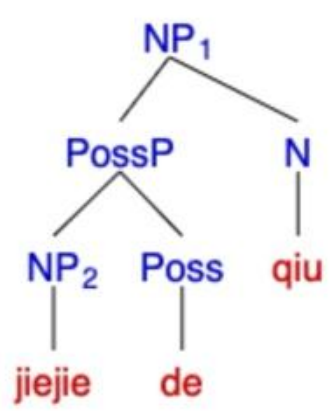

**Figure 6 Dropping Interpretation 1.**

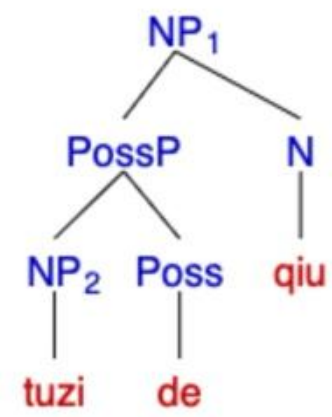

**Figure 7 Dropping Interpretation 1.**

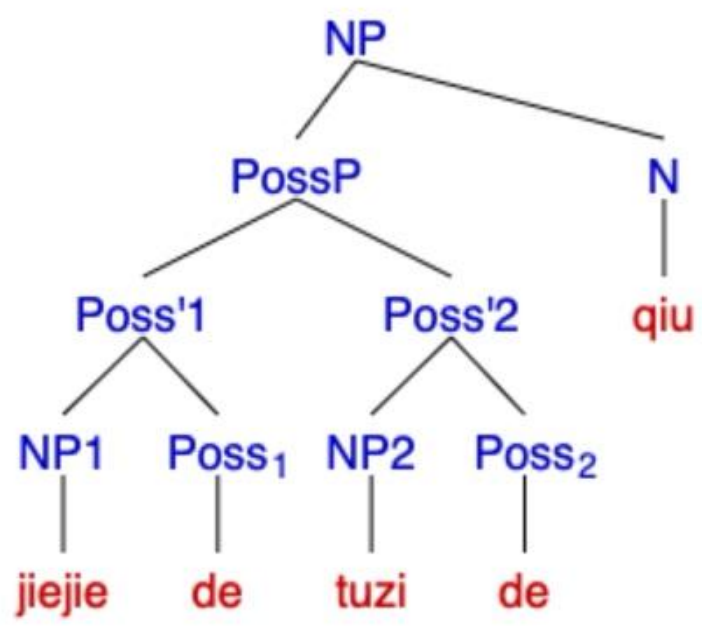

**Figure 8 Conjunctive Interpretation.**

In the realm of recursive location, the conjunction error constituted 7.82%, and the drop error made up 14.40% of all non-target responses. Notably, the drop error emerged as the predominant type of misinterpretation (11.73%), with the conjunction error also figuring prominently as a primary mode of incorrect interpretation (7.25%). The "drop" error still remained the predominant mode of incorrect response, with 3-year-olds exhibiting this error in 20.67 percent of their replies, a figure which trended downward to 11.52 percent among their 6-year-old counterparts. This decline suggests an age-related improvement in the children's language processing abilities. Similarly, the incidence of "conjunction" errors, where subjects incorrectly conjoin elements, also increased with age, reaching its zenith at 10.34 percent in the responses of 5-year-olds. This peak in error rates at this age may indicate a developmental stage where children are more prone to attempt complex linguistic structures, potentially leading to increased mistakes in conjunction. Figure 9 reflects the distribution of incorrect answer type by participants at every age level. This analysis underscores a developmental trend in children's comprehension of complex recursive structures, demonstrating an initial propensity for simplification errors that abates with increasing cognitive sophistication associated with age.

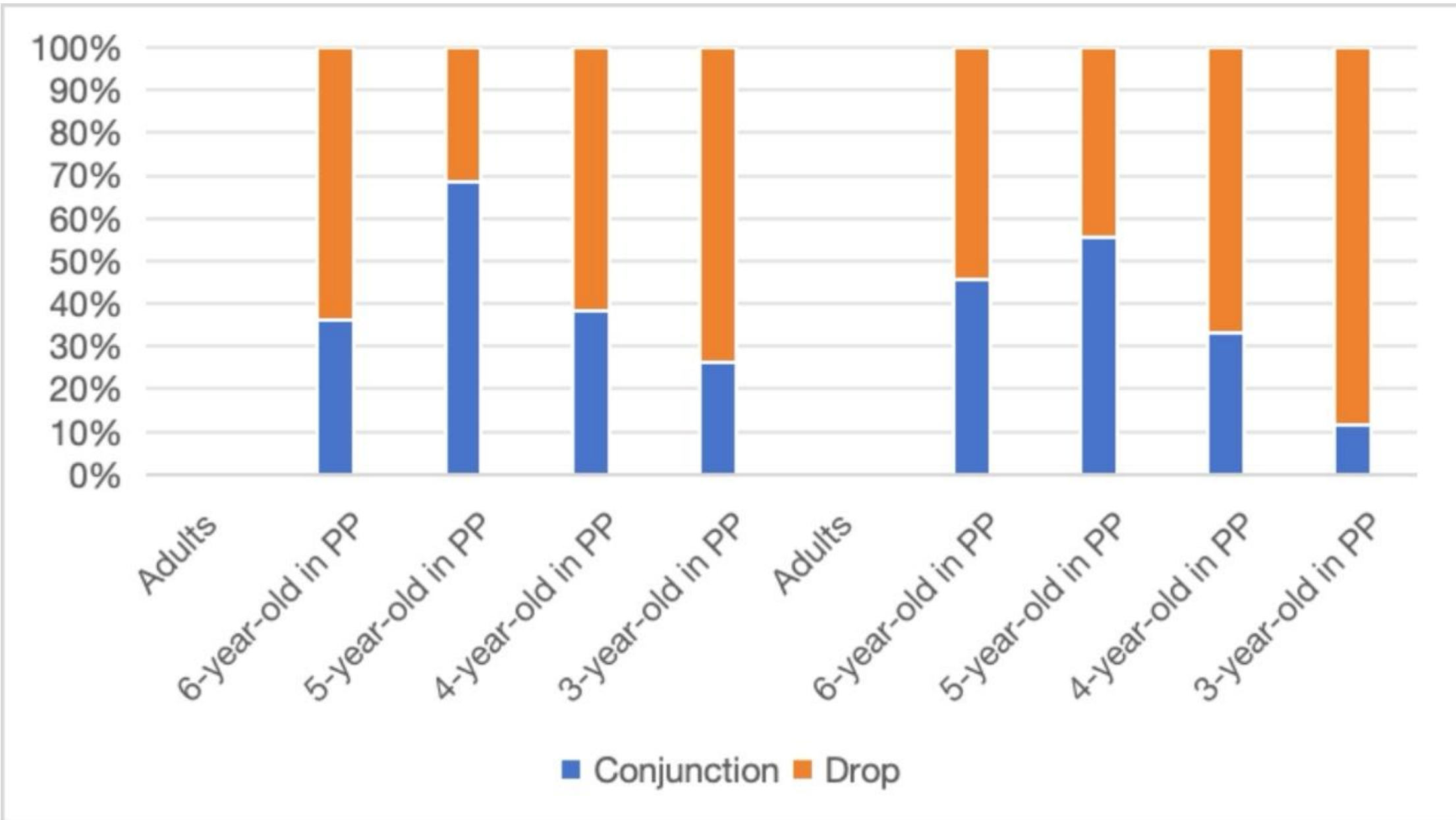

**Figure 9 The Error Distribution of Conjunction and Dropping.**

We then examine the difference between age and recursive possession and difference between age and recursive location. Specifically, based on the mean accuracy score of different types (Table 2) and standard derivation, so as to shed light on characteristics in the process in which Mandarin children acquire recursive possessives, further analysis of the data was conducted in SPSS. Repeated-measures ANOVA test was performed to test the correlations between potential factors. The Repeated-measures ANOVA test which set Types as within-subject factor and Age as the between-subject factor was carried out with proportion of accurate responses to questions as the dependent measure to determine whether there was significant interaction between Age Group and Recursive Types. The result demonstrate that age and types reflect significant effect for scores (F=8.399, p=0.000, df=4), revealing that recursive types have an effect on the accuracy.

| | Recursive Possession | | Recursive Location | |
|---|---|---|---|---|
| Age | Mean | S.D. | Mean | S.D. |
| 3 years(n=30) | 2.60 | 1.22 | 2.17 | 1.085 |
| 4 years(n=34) | 2.71 | 1.15 | 1.91 | 1.357 |
| 5 years(n=28) | 3.00 | 1.09 | 2.21 | 1.371 |
| 6 years(n=34) | 3.50 | 0.56 | 2.59 | 1.373 |
| Adults(n=30) | 3.97 | 0.18 | 3.83 | .379 |
| Total | 3.16 | 0.84 | 2.53 | 2.53 |

**Table 2. The mean value and standard derivation of the accuracy of different recursive types.**

Additionally, Kruskal-Wallis Test one-way analysis of Pairwise Comparison was run to compare the recursion interpretation of children and adults in those two types. Because none of the three data sets are normally distributed, Kruskal-Wallis Test one-way analysis of Pairwise Comparison of Age which is processed for the acquisition age of structures with different recursive types can present the following facts. In the recursive possessive structure, there were significant differences in correctness between the 3-year-old group, the 4- and 5-year-old groups and adult group (p=0.000, 0.000, 0.001 respectively), while there was no significant difference between the 6-year-old group and the adult group (p=0.09>0.05), suggesting that the recursive possessive structure had been acquired by the children at age of 6, whereas children younger than 6 did not acquire this type of structure. In recursive locative structure, there were significant differences between the 3-year-old group, the 4-, 5-, and 6-year-old groups and the adult group (p=0.000 for all), indicating that the recursive locative structure was not acquired by the children even at the age of 6.

The differences that emerge in the acquisition process among children of different ages is also a striking research feature. Language acquisition is not an overnight process, but a gradual one. We conducted a comparative analysis on each group of participants to better observe the acquisition of children across all ages. The paired comparison results in the Kruskal-Wallis Test display that in recursive possessive structure, there is no significant difference between 3-year-old and 4-year-old and 5-year-old group (p=1.000 for both group), but there is a significant difference between 3-year-old group and 6-year-old group ($p=0.04<0.05$). There is no significant difference between 4-year-old and 5-year-old and 6-year-old groups (p=1.000, 0.06, respectively) and no significant difference between 5-year-old and 6-year-old groups (p=1.000). In recursive locative structure, there is no significant difference between 3-year-old and 4-year-old, 5-year-old, 6-year-old groups (p=1.000, for all), and there is no significant difference between 4-year-old and 5-year-old groups and 6-year-old groups (p=1.000, 0.28, respectively), and there is no significant difference between 5-year-old and 6-year-old groups (p=1.000). The experimental data show that there is no significant difference between the two adjacent age groups, except between 3 - and 6-year-olds in recursive possession. We can find from the analyzed data that children 1 year apart have no significant difference in the acquisition of recursive possessive structure, but the significance increases with the increase of age gap. Combined with the above analysis of adult and child data, six years old is a crucial time node for the acquisition of recursive possessive structure. More importantly, this finding clearly shows that children's linguistic output matures bit by bit and that language, in essence, is "an organ of the body, a mental organ" (Berwick & Chomsky, 2016, p. 56).

Finally, there is a correlation between the acquisition of recursive possession and recursive location. There is a significant high positive relationship among the structure of recursive possession and recursive location ($p<0.05$), where recursive possession and recursive location are lowly correlated, $\rho=0.302$. This means that as the score of recursive possession increases, so does the score of recursive location, although the trend is not very strong.

## 6. Discussion

The results of recursive possessives show that Mandarin-speaking children are capable of understanding the structure for the two-level recursion of possessive phrases by the age of 6. However, Mandarin children still have a significant difference with adults in their understanding of recursive possession and recursive location until the age of six. This is consistent with the results of some previous studies (Matthei 1982; Roeper 2007, 2011; Pérez-Leroux et al. 2012；Limbach et al., 2010; Author et al., 2022; Author, 2014; Roberge et al. 2018), in which Children are not able to properly understand or produce two-level or higher recursive structures until a relatively late stage of language development. In the acquisition of recursive location, Our experiment found that Mandarin-speaking children had not learned recursive location until they were six years old, which is in contrast with Mao et al.'s (2024) finding. In their study, children are capable of generating two-level recursive location at the age of four in pointing task, which contributes to the ambiguous research method.

Despite evidence that four-year-olds can comprehend and produce one-level recursive constructions (Author et al., 2023), the current study indicates that children do not grasp two-level recursion until they reach the age of six. As two-level recursion is considered the true hallmark of recursion, the mastery of this level is indicative of a child's full recursion capability (Roeper, 2011; Martins & Fitch, 2014). The experimental findings of this paper thus suggest a relatively late acquisition of recursion skills in children.

Besides, the acquisition of recursion is a gradual process in Mandarin, consistent with what has been reported for Chinese (Author et al. 2022, 2023; Mao et al. 2024) and other languages like English (Pérez-Leroux et al., 2012; Iain Giblin et al. 2019) and Japanese (Roeper et al. 2012; Terunuma et al. 2018). Age is a very important factor in the acquisition of recursive possessives and locatives in Chinese-speaking children, influencing children's language development. Specifically, adults tend to understand the two structures more easily for they gave the correct answers for 99.17% and 98.67% respectively for one chance, whereas children are still limited in understanding target structures, who might need the second time to answer correctly. This finding is basically in line with the findings in Friederici et al. (2017) that confirms child language faculty and its corresponding brain areas are still in development.

In addition, different types of recursion also have an impact on acquisition. Our results suggest that there is a developmental stage where the structure for the recursion of possessive phrases is available but the structure for recursion of locative phrases is not. The comparison between the results of locatives with those of possessives demonstrates that there is some asymmetry between recursive possessives and recursive locatives concerning when children move on from one developmental stage to the next. One language internal factor assigned a determining role in timing of acquisition is complexity (Pérez-Leroux et al., 2022). There exists difference in acquiring different types of tail- recursion. In Mandarin-speaking children, the recursion of possessive phrases is available at around the age of 6, whereas that of locative phrases is not. Our results also show that children's performance in recursive locative sentences is generally worse than that in recursive possessive sentences at each age level. It can be said that recursive locatives are more difficult for children to comprehend than recursive possessives in general. The result is consistent with previous study (Pérez-Leroux et al., 2012; Terunuma et al. 2018; Pérez-Leroux et al., 2022) which argued that recursion is not acquired simultaneously for all types of modifiers. Children would come to understand recursive possession in an adult-like manner at the age of 6, but the ability to understand recursive locative structure at the age of six is still quite different from that of adults. There are possible reasons for this asymmetry.

Initially, children acquire recursive sequences across various degree of mergers at distinct developmental stages, with each subclass presenting unique challenges in the acquisition of recursion (Author et al., 2023). The progression to two-level recursive structures is not symmetrical among children. Further experimental data indicate that while two-level recursive possessive constructions are mastered by the age of 6, the acquisition of two-level recursive locative constructions, which are also a form of tail recursion, lags behind. This disparity can be attributed to the degree of merger in two kinds of two-level tail recursion, resulting in a significant variation in the number of merges required. In this study, when dealing with the complex structure of recursion, compared to possessive recursion, locative recursion also requires searching for specific grammatical elements such as locative words during the construction of complex hierarchical structures, searching and matching according to the syntactic features of different words, thereby activating the supramarginal gyrus (L. SMG). Specifically, two-level recursive possessives involve fewer merges than two-level recursive locatives, which may account for the earlier acquisition of the former. The observed asymmetry can be explained by the higher degree of merger associated with recursive locatives compared to recursive possessives (Fukui, 2017). The neurolinguistic research conducted by Ohta et al. (2013), Fukui (2017), Friederici (2014, 2017), and Friederici et al. (2017) has substantially substantiated the aforementioned conclusions. Ohta et al. (2013) demonstrated that both the Degree of Merger (DoM) model and the DoM combined with the number of search models were instrumental in explaining the neural activation within the left inferior frontal gyrus (L. F3op/F3t) and the supramarginal gyrus (L. SMG). Activation in the left inferior frontal gyrus (L. F3op/F3t) was directly correlated with DoM, signifying heightened activation in response to syntactically complex sentences, particularly those involving recursion. Meanwhile, supramarginal gyrus (L. SMG) activation was associated with both DoM and the "number of Search," suggesting its role in the search and application of syntactic features within complex sentence structures. Fukui (2017), through an fMRI experiment, determined that the processing of syntactically complex sentences, as quantified by the Degree of Merger, was positively correlated with the extent of neural activity in the left frontal operculum region (F3op/F3t), with more complex sentences eliciting greater neural responses. Furthermore, the processing of syntactic recursion at varying levels was also positively correlated with the neural activity in the left supramarginal gyrus (SMG), with one-level recursion exhibiting less activity compared to the more extensive activation observed with two-level recursion. Consistent with Fukui's findings, Friederici et al. (2014) conducted an fMRI study with German adults on two-level recursive sentence structures, revealing activation in the left inferior frontal gyrus, the posterior superior temporal gyrus, and the fiber bundles interconnecting these regions during the processing of recursion. Comparative analysis between adults and seven-year-old children revealed underdeveloped fiber bundles in children, impairing their ability to process two-level recursive structures. These collective neurolinguistic research achievements align with the conclusions presented in this paper: different recursive syntactic structures vary in complexity and are likely acquired independently at various developmental stages. This suggests a correlation between the "growth tree" of recursive syntactic development and the "growth tree" of neural development in the brain, as posited by Friedmann et al. (2021).

Secondly, a pivotal inquiry in the acquisition of Mandarin concerns how children discern whether a grammatical rule is capable of recursion, particularly in the realms of possession and location (Roeper & Snyder 2004; Roeper 2007, 2011). During the acquisition of Mandarin, children typically grasp the employment of the particle "*de*" to denote possession and locative relations at approximately 2 and 3 years of age, respectively (Shi et al., 2018; Mao et al., 2024). It is plausible that children initially internalize these usages as constrained lexical rules. Subsequently, children progress to a stage where they extend these rules and attempt to apply them to additional categories. In Mandarin, this generalization is largely permissible across the majority of nouns. This trend is particularly evident among younger children aged 4 to 6 and preadolescents aged 10 to 12, indicating distinct developmental trajectories of overgeneralization and subsequent refinement. Conversely, in the context of locative expressions, children exhibit a reluctance to overgeneralize the use of the locative marker. This reluctance is attributed to the locative marker's restricted distribution, which does not readily lend itself to broader application. The selective application of the locative marker by children underscores the nuanced developmental processes underlying the acquisition of recursion in Mandarin, where the generalization of grammatical rules is not uniform across different semantic domains.

Furthermore, the propensity for utilizing certain notional domains to contrastively identify referents appears to be influenced more by extralinguistic factors than by language-specific elements. Culbertson et al. (2020: 696) posit that the concept of possession is more readily employed for contrastive reference than location, suggesting that the linguistic patterns associated with possession are influenced by cognitive features. This perspective is reinforced by the findings of Pérez-Leroux et al. (2022), which demonstrate that German-speaking participants exhibit greater proficiency in recursive possessive constructions compared to locative constructions. This phenomenon is not isolated; it is observed consistently across age groups and languages, as similar asymmetries in notional domain usage have been documented in studies on French, English, and Japanese (Robergeet al., 2018； Pérez-Leroux et al., 2022; Terunuma et al., 2018). We thus hypothesize that, cross-linguistically, possessive notional domains are more likely to be employed for contrastive reference than locative relationships. This inclination is attributed to inherent cognitive characteristics. The preference is not merely for specific syntactic configurations but is instead determined by the degree of relational strength between the head noun and the modifier within a notional domain. This relational strength is what renders a particular notional domain more or less susceptible to being used for contrastive reference, thereby yielding the observed patterns.

Theoretically, the developmental trajectory of recursion in preschool-aged children is characterized by incremental stages. 4-year-olds are considered to be in a nascent phase of recursion ability, possessing the capacity for one-level recursion, yet exhibiting instability in their performance (Author, 2014). By the age of five, children's recursive skills are more advanced, and by six years old, their abilities have generally reached a plateau, approximating the proficiency observed in adults, with minimal variability. This progression suggests a gradual maturation of children's understanding and production of recursive language structures. Additionally, the study's results, which include responses that diverge from the expected answers, underscore that children under the age of six have not yet internalized the two-level recursive structures. This difficulty in understanding is further evidenced by their inability to process two-level recursion with the ease observed in adults.

The majority of the children's responses were structured around the topic but did not conform to the anticipated syntactic structure. The responses can be categorized into three types: conjunction, dropping, and other interpretations.

The first type of response involves the division of the two-level recursive structure according to a leftward branching direction, combining the input structure of the first two layers and then coordinating it with the third layer. The second type of response entails the omission of one of the recursive layers in their understanding, as seen in utterances such as "the boy's balloon" or "the dad's balloon." The third type of response includes the color of all the key words in the description. In conclusion, the acquisition of two-level tail recursion is relatively late, occurring at around the age of six, and prior to this, children encounter significant challenges in comprehending two-level internal recursive structures.

Concerning the factor that may influence children’s acquisition of recursive possession, age is the pivotal factor. According to Martins and Fitch (2014), a child's recursion ability is ascertained by their capacity to innovatively acquire two-level recursive structures that they have not previously encountered, based on their understanding of one-level internal recursion. The findings of this study indicate that while children rapidly develop the ability to comprehend one-level recursive structures, they do not exhibit the creative capacity to grasp two-level recursion until around the age of 6. During the interval between ages 3 to 6, there is a notable increase in the number of children achieving two-level recursion proficiency. This developmental trajectory of internal recursion aligns with the tenets of the Maturation Theory (Wexler 1990, 2004), which posits that children's linguistic faculties, akin to other physiological systems, undergo a specific maturational process. In the maturational approach, development is determined primarily by internal factors that are controlled by genes. Developmental change is assumed to be based solely on a maturational blueprint; the actual sequence is invariant, but the rate is variable (Aylward, 2020). According to this perspective, recursion is an innate faculty that necessitates a gradual developmental trajectory and unfolds naturally at a predetermined stage, analogous to the natural timing of physiological milestones such as the eruption of permanent teeth around the age of 6. It is also the perspective that underlies much of the traditional, often rudimentary material, recursion, on child development that was presented in school and training, where the focus was on milestones and norms being on time or delayed. In this model, development depends entirely on neurological and physical maturation, and it proceeds in fixed sequences. The child is seen as an immature or incomplete organism that moves in predictable patterns of behavior during the course of continuous maturation. It is a linear model, with development seen as quantitative gains in competencies. Although these milestones and age norms have been modified slightly over time, the sequences still ring true. They form the core for most developmental tests currently in use. Classification of children as delayed, deviant or normal based on varying rates of emergence of specific skills follows from these observations (Dixon, 2006). The developmental pattern of children's internal recursion abilities, as revealed in this study, is most coherently accounted for within the framework of the Recursion Only Hypothesis (ROH). This hypothesis posits that the acquisition of recursive abilities is a natural endowment that unfolds in accordance with a child's maturation, serving as a pivotal piece of evidence in the discourse on the language acquisition device.

## 7. Conclusion

This empirical study presents an in-depth analysis of the acquisition of recursive possessive and locative constructions among Mandarin-speaking children. The research scrutinizes the developmental stages in children's comprehension of two-level recursive structures, revealing that full mastery of adult-like recursion is typically not attained until the age of 6. The study also underscores the significance of notional domains in the acquisition process, suggesting that children's facility with possessive and locative recursion is shaped by the underlying conceptual frameworks. These findings contribute meaningfully to the theoretical discourse on language development, providing valuable insights into the interplay between universal grammatical principles and the acquisition of language-specific structural nuances in early childhood. The research enriches our understanding of the cognitive mechanisms underlying language acquisition and the role of recursion in children's linguistic development.

**Data Availability Statement**

The data supporting the findings of this study are available to reviewers and interested researchers. The anonymized dataset has been uploaded to *OSF* and can be accessed via the following link: https://osf.io/98up4/?view_only=68815337b6f34458b4c1f9f0ec98ec39. For further information regarding data access or any inquiries related to the study, please contact the corresponding author at 2532872730@qq.com. The authors confirm that all data and materials associated with this study comply with the principles of transparency and reproducibility as required by the journal's guidelines.

**Ethics and Consent Statement**

This study was conducted in full compliance with the ethical standards as outlined in the 1964 Declaration of Helsinki and its subsequent amendments. The research protocol was reviewed and approved by the Institutional Review Board for the Protection of Human Subjects at Suzhou University, Jiangsu Province, China. Written informed consent was obtained from the legal guardians of all child participants involved in the study. The confidentiality and privacy of the participants were strictly maintained throughout the research process. All data collected were anonymized and no identifying details were included in the manuscript and supplementary materials to ensure the protection of the privacy and anonymity of the participants.